\documentclass[review]{elsarticle}

\usepackage{amssymb}
\usepackage{dsfont}
\usepackage{bbm}
\usepackage{amsfonts}
\usepackage{color}
\usepackage{multicol}
\usepackage{graphicx}
\usepackage{setspace}
\usepackage{amsmath}
\usepackage{amssymb}
\usepackage[pdftex,linkcolor=blue,citecolor=blue,backref=page]{hyperref}

\journal{Journal of Neurocomputing}

\begin{document}

\begin{frontmatter}

\title{3G Structure for Image Caption Generation}




\author[a,b]{Aihong Yuan}
\author[a]{Xuelong Li}
\author[a]{Xiaoqiang Lu\corref{c}}
\cortext[c]{Corresponding author}
\ead{luxq666666@gmail.com}

\address[a]{Center for OPTical IMagery Analysis and Learning (OPTIMAL),
  \\Xi'an Institute of Optics and Precision Mechanics,
  Chinese Academy of Sciences, \\Xi'an 710119, Shaanxi, P. R. China.}
\address[b]{University of Chinese Academy of Sciences, Beijing 100049, P. R. China.}

\begin{abstract}\footnote{2019 Elsevier. Personal use of this material is permitted. Permission from Elsevier must be obtained for all other uses, in any current or future media, including reprinting/republishing this material for advertising or promotional purposes, creating new collective works, for resale or redistribution to servers or lists, or reuse of any copyrighted component of this work in other works.}
 It is a big challenge of computer vision to make machine automatically describe the content of an image with a natural language sentence. Previous works have made great progress on this task, but they only use the global or local image feature, which may lose some important subtle or global information of an image. In this paper, we propose a model with 3-gated model which fuses the global and local image features together for the task of image caption generation. The model mainly has three gated structures. 1) Gate for the global image feature, which can adaptively decide when and how much the global image feature should be imported into the sentence generator. 2) The gated \textsl{recurrent neural network} (RNN) is used as the sentence generator. 3) The gated feedback method for stacking RNN is employed to increase the capability of nonlinearity fitting. More specially, the global and local image features are combined together in this paper, which makes full use of the image information. The global image feature is controlled by the first gate and the local image feature is selected by the attention mechanism. With the latter two gates, the relationship between image and text can be well explored, which improves the performance of the language part as well as the multi-modal embedding part. Experimental results show that our proposed method outperforms the state-of-the-art for image caption generation.
\end{abstract}


\end{frontmatter}


\section{Introduction}
\label{Introduction}
Image caption generation aims to automatically generate a natural language sentence to describe the content of a given image. It is a vital task of scene understanding which is one of the fundamental goals of computer vision and artificial intelligence \cite{ DBLP:conf/eccv/AhmadC16, DBLP:conf/cvpr/DhimanTCC16}. However, image caption generation is a challenging task. It not only needs to recognize objects in an image, but also needs to capture and express their relationships and attributes with natural language \cite{vinyals2015nic}.
\par To address the aforementioned challenge, many methods have been developed and a lot of gratifying results have been achieved in recent years. These methods are roughly divided into two categories: 1) retrieval-based methods \cite{hodosh2013framing, kulkarni2013baby, socher2014dt-rnn} and 2) \textsl{multi-modal neural network} (MMNN) based methods \cite{DBLP:journals/pami/VinyalsTBE17, DBLP:journals/pami/KarpathyF17, 7792748, 7558228}. Although these methods, especially the MMNN-based methods, have attained promising results, further improvements should be got over some limitations.

\subsection{Motivation and Overview}
To generate length-variable and form-variable  sentences, we follow the MMNN-based methods. However, in most existing methods, only global image feature or local image feature is used to generate sentences, which may lose some important image information. It is easy to understand that global image feature can catch the overall information of an image and local image information can dig out the fine-grained relationships between image regions and language elements. Therefore, we use a global-local image feature fusing strategy, which can fully mine the image information.
\par Moreover, the sentence generator in image captioning model should learn both hierarchical and temporal representation very well. While, most of the previous methods use the single-layer ``vanilla'' RNN \cite{mao2014m-rnn, karpathy2015devs, DBLP:journals/pami/KarpathyF17} or single-layer LSTM, which cannot learn the hierarchical representation well \cite{DBLP:conf/icml/ChungGCB15, DBLP:journals/corr/ChungAB16}. Meanwhile, the issue of learning multiple adaptive timescales (\textsl{i.e.} the quickly and slowly changing components) \cite{DBLP:conf/icml/ChungGCB15, DBLP:journals/corr/ChungAB16} should also be considered in caption generation, because the sentence is a sequence signal which consists of both fast-moving and slow-moving components \cite{DBLP:conf/icml/ChungGCB15}. In the natural language processing field, \emph{gated feedback RNN} (GF-RNN) has attracted the attention of many researchers, because GF-RNN can not only solve the long-term dependency problem in ``vanilla'' RNN, but also learn multiple adaptive timescales.

\par Motivated by the aforementioned reasons, we propose a novel 3G model in this paper. More specially, 1) both the global and local image features are input into the multi-modal embedding part for making full use of image information; 2) the gated RNN and gated feedback connecting strategy are used to improve the performance of the language model.
This model mainly contains 3 gated modules: \textbf{gate} for global image feature, \textbf{gated RNN} and \textbf{gated feedback connecting} for stacked RNN. So we name it 3G for short.
When we fuse the global and local image feature, the first gate is used to control when and how much the global image information should be input into the multi-modal embedding part. The gate is motivated by the gate structure of \textsl{long-short term memory} (LSTM) unit.
Recently, RNN is used as the most popular language model. However, ``vanilla'' RNN is hard to capture the long term dependencies. So we choose LSTM, one kind of gated RNN, as the language model. It is the second gate. Moreover, in order to explore the nonlinear relationship between the image and text, the most effective strategy for stacked RNN---\textsl{gated feedback RNN} (GF-RNN) \cite{DBLP:conf/icml/ChungGCB15} is used as the multi-modal part as well as the sentence generator. It is the third gate. With the 3-gated structure, the image information is fully utilized and the performance of the language model is strengthened.

\subsection{Contributions}
\par Our main contributions are listed as follows:
\begin{enumerate}
  \item An end-to-end 3G model is proposed to accomplish the image caption generation task. The proposed model contains 3-gated structure, and it can be fully trained with the \textsl{stochastic gradient descent} (SGD) method.
  \item The global and local image features are used in this paper, which makes full use of image information to improve the caption quality.
  \item The gated feedback connecting strategy is used for stacking the LSTM, which solves the issues of long-term dependency and learning multiple adaptive timescales. In other words, the gated RNN and gated feedback connecting strategy make the multi-modal embedding and language parts much stronger than the previous methods.
\end{enumerate}

A shorter version of this paper appears in \cite{DBLP:conf/cccv/YuanLL17}.  The main extensions in the current work are:
\begin{enumerate}
  \item In terms of innovation, we have added a new technological novelty in the caption generating part: gated feedback LSTM is first used to generate the description for image and the gated feedback connecting strategy can well explore the nonlinear relationship between text and image.
  \item Since the language part in this paper is GF-LSTM, the formulas of the GF-LSTM is introduced in detail in Section \ref{GF-RNNs} and the language generating process modeled by the GF-LSTM is detail stated in Section \ref{our model}.
  \item More experiments and more experimental details has been added in Section \ref{Experiments}.
\end{enumerate}

\subsection{Organization}
\par The rest of this paper is organized as follows. In Section \ref{Related Work}, some previous works are briefly introduced. Then the \textsl{gated feedback LSTM} (GF-LSTM) model is introduced in Section \ref{GF-RNNs}. Section \ref{our model} presents our model for image caption generation. To validate the proposed method, the experimental results are shown in Section \ref{Experiments}. At last, Section \ref{conclusion} makes a brief conclusion for this paper.

\section{Related Work} \label{Related Work}
The problem of describing images with natural language sentences has recently attracted increasing interests and many methods have been proposed. These methods can be roughly divided into two categories: Retrieval-based Methods \cite{hodosh2013framing, kulkarni2013baby, socher2014dt-rnn} and MMNN-based methods \cite{DBLP:journals/pami/VinyalsTBE17, DBLP:journals/pami/KarpathyF17}.

\begin{figure}[ht]
  \centering
  \includegraphics[width=0.8\linewidth]{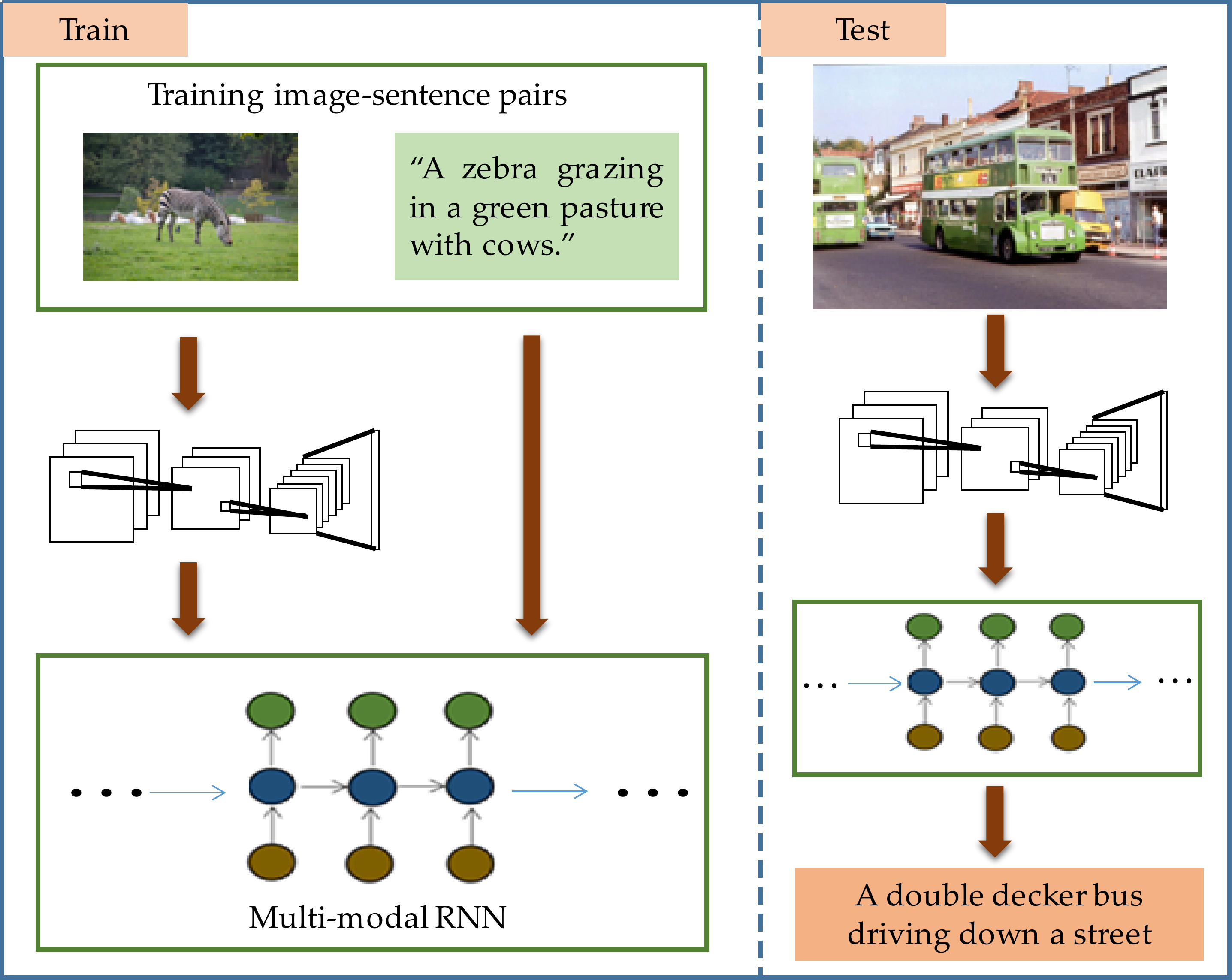}\\
  \caption{The most popular paradigm for image caption generation. In this diagram, CNN and RNN are used as visual feature extractor and sentence generator, respectively. In the training step (the subfigure on the left of the dotted  line), the image feature and the corresponding sentence representation are imported into the multi-modal RNN to learn the mapping relation between image and text. Then in the testing step (the subfigure on the right of the dotted line), only the image feature is imported into RNN, and then the image caption is generated by RNN.}
  \label{illustration_fig}
\end{figure}

\par \textsl{Retrieval-based methods} retrieve similar captioned images and generate new descriptions by retrieving a similar sentence from the training dataset \cite{hodosh2013framing}.  However, the style of the describing sentences generated by these kind methods is lack of variety.
\par With the wide use of \textsl{convolutional neural networks} (CNNs) and \textsl{recurrent neural networks} (RNNs) in computer vision and natural language processing \cite{7335630, szegedy2015googlenet, simonyan2014vggnet, krizhevsky2012alexnet, DBLP:conf/cvpr/HeZRS16, DBLP:journals/pami/DongLHT16, DBLP:journals/pami/LiuSL016, DBLP:journals/corr/Lipton15, chung2014empirical, cho2014gru}, \textsl{MMNN-based Methods} have become the most popular mechanism for image caption generation \cite{mao2014m-rnn, donahue2015lrcn, vinyals2015nic, 7546397}. As illustrated in Fig. \ref{illustration_fig}, the most common model contains three important parts: a vision part, a language module part and a multi-modal part.

\par MMNN-based methods can generate length-variable sentences and solve the drawbacks of retrieval-based methods. For instance, Mao \textsl{et al.} \cite{mao2014m-rnn} proposed a model called m-RNN to predict the next word conditioned on both previous words and the global image feature generated by CNN at each time-step.
Kiros \textsl{et al.} \cite{DBLP:journals/corr/KirosSZ14} proposed a similar joint multi-modal embedding model by using a powerful CNN and a LSTM that encodes text.
Vinyals \textsl{et al.} \cite{vinyals2015nic, DBLP:journals/pami/VinyalsTBE17} combined CNN with LSTM to create an end-to-end network that can generate natural language sentences for images.
Similarly, Karpathy \textsl{et al.} \cite{karpathy2015devs, DBLP:journals/pami/KarpathyF17} also proposed a multi-modal RNN model, and unlike \cite{mao2014m-rnn}, they use the global image feature generated by CNN only at the first time-step.
Therefore, one problem is coming: when and how much the global image information should be input into the RNN. These models do not reach a consensus and cannot solve this problem. To address this problem, gating mechanism is used to control the global image feature when and how much be input into the RNN in this paper.
The aforementioned models only use the global image feature output by CNN as the whole image information. However, they cannot provide fine-grained modeling of the inter-dependencies between different visual elements and the relationship between the image and text \cite{7792748}. That is the second problem
\par To address the second problem, attention-based methods has been proposed \cite{7792748, DBLP:conf/icml/XuBKCCSZB15, DBLP:conf/cvpr/YouJWFL16, DBLP:conf/nips/YangYWCS16}.
Xu \textsl{et al.} \cite{DBLP:conf/icml/XuBKCCSZB15} exploited attention mechanism for image caption generation. It used the feature map output from the convolutional layer of the CNN as image information. By flattening the feature map into a fixed number vectors, every image was broken into tiles with fixed size. Each vector denotes one tile feature. When predicting the next word, the previous generated word will select the tiles. \cite{DBLP:conf/cvpr/YouJWFL16} generates natural language sentence with attention transitioning on the lexical representation.
Moreover, attention mechanism is widely used in other computer vision and natural language processing tasks \cite{DBLP:conf/nips/MnihHGK14, DBLP:journals/corr/BaMK14, DBLP:journals/corr/BazzaniLT16, DBLP:journals/corr/BahdanauCB14}.
\par In this paper, we retain the attention part as the local image feature. To improve the performance of the language part, gated feedback connecting strategy is used for stacking the LSTM.

\section{Gated Feedback LSTM} \label{GF-RNNs}
\begin{figure*}[t]
  \centering
  \includegraphics[width=0.95\linewidth]{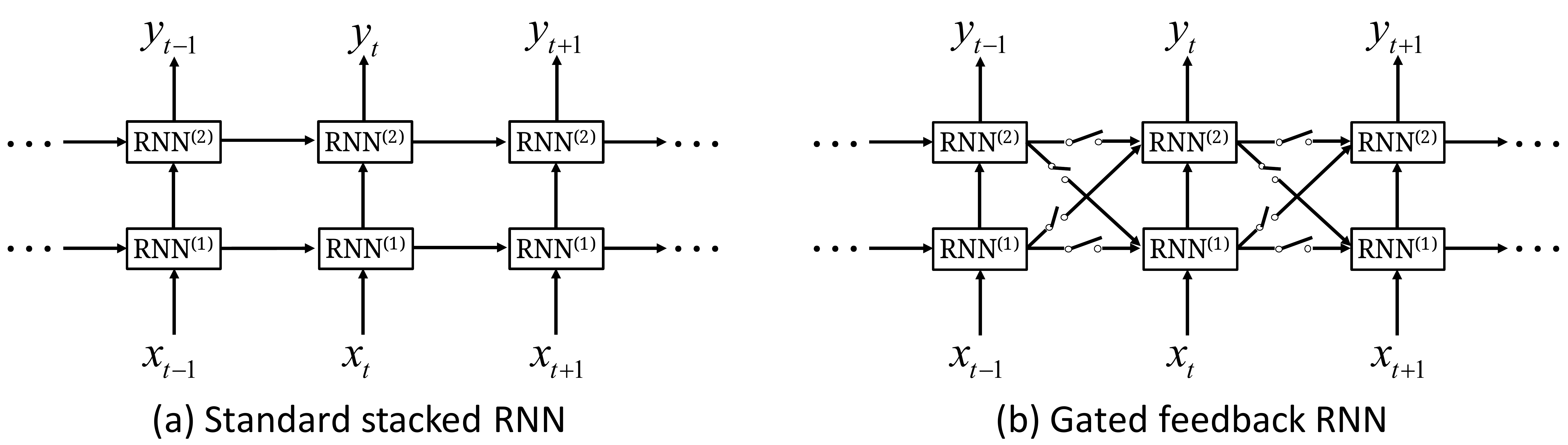}\\
  \caption{Two kinds of stacking strategy to form a deep RNN architecture: (a) standard stacked RNN and (b) gated feedback RNN. ``Switches'' in (b) denote gates which control whether the previous hidden states are used to compute the current states.}
  \label{GF-RNN}
\end{figure*}

In this paper, 2-layer GF-LSTM unit is used as sentence generator, so we introduce the GF-LSTM at first.
\par In previous works, sentence generation models based on RNNs have shown powerful capabilities to generate target sentences. In order to further enhance representing capabilities of RNN, a conventional way is stacking multiple recurrent layers \cite{DBLP:conf/icml/ChungGCB15}. However, as shown in Fig. \ref{GF-RNN}.(a), signals in standard stacked RNN only flow from lower layers to upper recurrent layers. Fig. \ref{GF-RNN}.(b) shows gated feedback stacking for RNN. Signals not only can flow from previous lower layers to current upper layers, but also can flow from previous upper layers to current lower layers. \cite{DBLP:conf/icml/ChungGCB15} has confirmed that GF-RNN shows the best performance on natural language processing among the standard stacking RNN with the same layers and single layer gated RNN (\textsl{e.g.} LSTM or GRU).

\par We take Fig. \ref{GF-RNN}.(b) as an example and choose LSTM as RNN model, according to \cite{DBLP:conf/icml/ChungGCB15}, the global gates which control previous hidden states to current hidden states are defined as follows:
\begin{equation}\label{gate-defining}
\left\{ {\begin{array}{*{20}{l}}
{g_h^{i \to 1} = \sigma \left( {{\bf{w}}_{{g_h}}^{i \to 1} {{\bf{x}}_t} + {\bf{u}}_{{g_h}}^{i \to 1} {\bf{h}}_{t - 1}^{(1,2)}} \right)}\\
{g_h^{i \to 2} = \sigma \left( {{\bf{w}}_{{g_h}}^{i \to 2}  {\bf{h}}_t^{(1)} + {\bf{u}}_{{g_h}}^{i \to 2}  {\bf{h}}_{t - 1}^{(1,2)}} \right)}
\end{array}} \right.,
\end{equation}
where superscript $i = 1, 2$ denotes the level of layers, superscript $i \to j$ $(j = 1,\ 2)$ denotes state transiting from layer $i$ to layer $j$. ${\bf{w}}_{{g_h}}^{i \to j}\in \mathbb{R}^{1\times h}$ stands for weight vector for the current input and ${\bf{u}}_{{g_h}}^{i \to j}\in \mathbb{R}^{1\times 2h}$ $(j = 1,\ 2)$ represents weight vector for the previous hidden states. $
{\bf{h}}_{t - 1}^{(1,2)} = {\left[ {\begin{array}{*{20}{c}}
{{{\left( {{\bf{h}}_{t - 1}^{(1)}} \right)}^T}}&{{{\left( {{\bf{h}}_{t - 1}^{(2)}} \right)}^T}}
\end{array}} \right]^T} \in \mathbb{R}^{2h}$ denotes the previous states. Through Eq. (\ref{gate-defining}) we know that gate $g_h^{i \to j}$ is a single scalar and its value depends on the current lower hidden state and the previous hidden states.

\par After computing the gates, we describe how to use these gates in LSTM. When computing the current input gate, output gate, forget gate, memory and the current hidden state, $g_h^{i \to j}$ is not used, so these formulas are the same as raw LSTM formulas. We rewrite the three gate formula of LSTM unit as follows:
\[{\bf{i}}_t^{(i)} = \sigma \left( {{W_i^{(i)}}{\bf{h}}_t^{(i - 1)} + {U_i^{(i)}}{\bf{h}}_{t-1}^{(i)} + {\bf{b}}_i^{(i)}} \right),\]
\[{\bf{f}}_t^{(i)} = \sigma \left( {{W_f^{(i)}}{\bf{h}}_t^{(i - 1)} + {U_f^{(i)}}{\bf{h}}_{t-1}^{(i)} + {\bf{b}}_f^{(i)}} \right),\]
\[{\bf{o}}_t^{(i)} = \sigma \left( {{W_o^{(i)}}{\bf{h}}_t^{(i - 1)} + {U_o^{(i)}}{\bf{h}}_{t-1}^{(i)} + {\bf{b}}_o^{(i)}} \right),\]
where superscript $i=1,\ 2$, ${W_{*}^{(i)}}\in {\mathbb{R}^{h\times h}}$ and ${U_{*}^{(i)}}\in {\mathbb{R}^{h\times h}}$ denote weights matrixes and ${{\bf{b}}_{*}^{(i)}}\in {\mathbb{R}^{h}}$ stands for the biases. ${\bf{i}}_t^{(i)}\in {\mathbb{R}^h}$, ${\bf{f}}_t^{(i)}\in {\mathbb{R}^h}$, ${\bf{o}}_t^{(i)}\in {\mathbb{R}^h}$ , ${\bf{h}}_t^{(i-1)}\in {\mathbb{R}^h}$ and ${\bf{h}}_{t-1}^{(i)}\in {\mathbb{R}^h}$ represent the current and the $i$-th layer's input gate, forget gate, output gate, the $i-1$-th hidden state and the previous $i$-th hidden state, respectively. When the superscript $i=1$, ${{\bf{h}}_t^{(i - 1)}}={{\bf{x}}_t}\in {\mathbb{R}^h}$. $\sigma ( \cdot )$ denotes the sigmoid activation function.
\par The current memory and the hidden state are computed as follows:
\begin{equation}\label{cell}
{\bf{c}}_t^{(i)} = {\bf{f}}_t^{(i)} \odot {\bf{c}}_{t - 1}^{(i)} + {\bf{i}}_t^{(i)} \odot {\bf{\tilde c}}_t^{(i)},
\end{equation}
\begin{equation}\label{hidden}
{\bf{h}}_t^{(i)} = {\bf{o}}_t^{(i)} \odot \tanh \left( {{\bf{c}}_t^{(i)}} \right),
\end{equation}
where ${\bf{c}}_t^{(i)}\in {\mathbb{R}^h}$ indicates the current memory, ${\bf{\tilde c}}_t^{(i)}\in {\mathbb{R}^h} $ denotes the updating memory content. ``$\odot$'' denotes element multiplication.
\par The computing formula of updating memory content is different from standard stacked LSTM. The updating memory content of the gated feedback LSTM at the $j$-th layer is computed as follows:
\[{\bf{\tilde c}}_t^{(j)} = \tanh \left( {W_c^{j - 1 \to j}{\bf{h}}_t^{(j - 1)} + \sum\limits_i {g_h^{i \to j}U_c^{i \to j}{\bf{h}}_{t - 1}^{(i)}} } \right),\]
where ${W_c^{j - 1 \to j}}\in {\mathbb{R}^{h\times h}}$ and ${U_c^{i \to j}}\in {\mathbb{R}^{h\times h}}$ stand for weights matrixes. Gate $g_h^{i \to j}$ is defined in Eq. (\ref{gate-defining}). Through this equation, we can know that $g_h^{i \to j}$ controls how much the previous hidden state ${\bf{h}}_{t - 1}^{(i)}$ is used to compute the current updating memory content.

\section{Proposed Approach} \label{our model}

\begin{figure*}[t]
  \centering
  \includegraphics[width=0.90\linewidth]{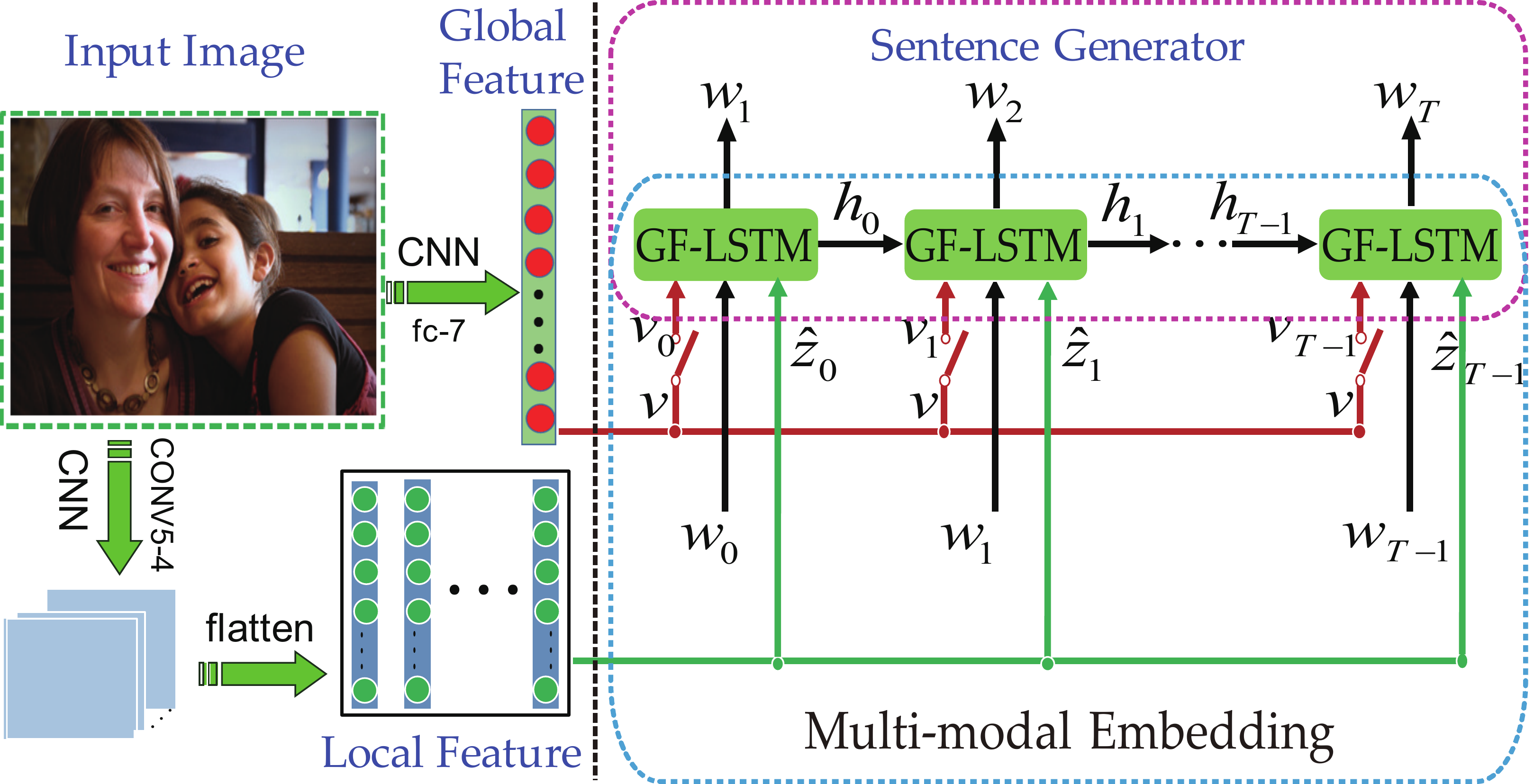}\\
  \caption{Overview of our method for image caption generation. A deep CNN for image features extraction: the FC7 layer is used to extract global features and the CONV5-4 layer is used to extract local features of the given images. RNN model acts as a decoder which decodes image features into sentences. In the training stage, image features and corresponding words vectors are input into the RNN with control. The gate for controlling the global image feature is computed with the pre-step hidden state of RNN. The local image features are selected by the pre-step hidden state and the current word vector.}
  \label{3G_fig}
\end{figure*}

\textbf{Overview.} Fig. \ref{3G_fig} shows the architecture of our 3G model. To make full use of the image information, the global and local image features are input into the multi-modal embedding module. The first gate controls when and how much the global feature is input into the multi-modal embedding module. While the local feature is selected by the attention mechanism. Another two gates are used in the language and the multi-modal embedding module parts. The gated structure is used to improve the performance of the language model. In this section, we interpret our 3G model in detail.

\par In the image caption generation task, when given an image, the most wanted sentence should be generated with a maximal probability. In many previous works, probability models are widely used. Similarly, our approach also uses probability generation models for this task. In other words, the LSTM module outputs a probability at each time-step. And we maximize the joint probability conditioned on the given image. So the objective function is written as follows:
\[P\left( {S\left| I \right.} \right) = \prod\limits_{i = 1}^N {P\left( {{S_i}\left| {{I_i};\theta } \right.} \right)}, \]
and we rewrite it as log likelihood function:
\begin{equation}\label{OF}
\log P\left( {S\left| I \right.} \right) = \sum\limits_{i = 1}^N {\log P\left( {{S_i}\left| {{I_i};\theta } \right.} \right)} .
\end{equation}
We compute $P\left( {{S_i}\left| {I_i} \right.} \right)$ with chain rule:
\begin{equation}
 P\left( {{S_i}\left| {I_i} \right.} \right) = \prod\limits_{t = 0}^{N_i} {P\left( {\left. {{w_t}} \right|I_i,{w_0}, \cdots ,{w_{t - 1}};\theta} \right)},
\end{equation}
where $\theta$ in these three equations denotes all of the parameters needed to train. $({I_i},{S_i})$ denotes the $i$-th image-sentence pairs and $N_i$ is the sentence length. In our model, GF-LSTM is used to model the condition probability (see Section \ref{GF-LSTM}).
\subsection{Image Representation}\label{Image Representation}
Almost all of the  state-of-the-art methods used deep CNN to encode the image, because deep CNN can learn discriminative and representative  features from the data such as the given images. Similar to the previous methods, we use the VGG-19 as the image features extractor. More specially, outputs of FC7 and CONV5-4 layers are used as global and local features of the images, respectively.

\subsubsection{Image Global Feature Representation} \label{gv}
The VGG-19 is pre-trained on ImageNet and used as the image encoder in our model. The global representation of image I is as follows:
\begin{equation}\label{global feature}
  {\bf{v}} = {{\bf{W}}_I}\cdot[Fc(I)] + {{\bf{b}}_I},
\end{equation}
where $I$ denotes the image I, $Fc(I)\in \mathbb{R}^{l}$ is the output of the FC7 layer. The matrix ${\bf{W}}_{I}\in \mathbb{R}^{h\times l}$ is an embedding matrix which projects $l$-dimension image feature vectors into the embedding space with $h$-dimension and ${\bf{b}}_{I}\in \mathbb{R}^{h}$ denotes the bias. ${\bf{v}}\in \mathbb{R}^{h}$ is the so-called image global feature representation.

\subsubsection{Image Local Feature Representation} \label{llv}
When a raw image I $\in {\mathbb{R}}^{W\times H\times 3 }$ is input to VGG-19, the CONV5-4 layer outputs feature map ${{\bf{v}}_c}\in \mathbb{R}^{W' \times H' \times D}$. Then, we flatten this feature map into ${{\bf{v}}_l}\in \mathbb{R}^{D\times C}$, where $C = W'\times H'$. This processing program can be written as follows:
\begin{equation}\label{local feature}
  {{\bf{v}}_l} = \left\{ {{{\bf{v}}_{l1}},{{\bf{v}}_{l2}}, \cdots ,{{\bf{v}}_{lC}}} \right\} = flatten\left( {Conv(I)} \right),
\end{equation}
where ${\bf{v}}_{li}\in \mathbb{R}^{D},\ i\in \{0,1,\cdots ,C\}$ denotes the feature of $i$-th location of image I. In other words, each image I is divided into $C$ regions and every ${\bf{v}}_{li}$ represents one region. In the training stage, the proposed method explores the relationship between words and image locations. In other words, when a word is imported, the word will guide which locations should be selected.
\subsection{Sentence Representation}\label{Sentence Representation}
In our model, we encode words into one-hot vectors.
We denote any sentence as $S=({{\bf{w}}_1}, {{\bf{w}}_2},\cdots ,{{\bf{w}}_N})$, where ${\bf{w}}_i\in \mathbb{R}^{N_0}$ denotes the $i$-th word in the sentence. We embed these words into embedding space. The concrete formula is as follows:
\begin{equation}\label{sentence representation}
  {{\bf{s}}_t} = {{\bf{W}}_s}\cdot {{\bf{w}}_t},\ t\in \{1,2,\cdots ,N\},
\end{equation}
where ${\bf{W}}_s$ is the embedding matrix of sentences which projects the word vector into the embedding space. So the projection matrix $\bf{W}_s$ is a $N_0\times h$ matrix where $N_0$ is the size of the dictionary and $h$ is the dimension of the embedding space.
\begin{figure*}[t]
  \centering
  \includegraphics[width=1.0\linewidth]{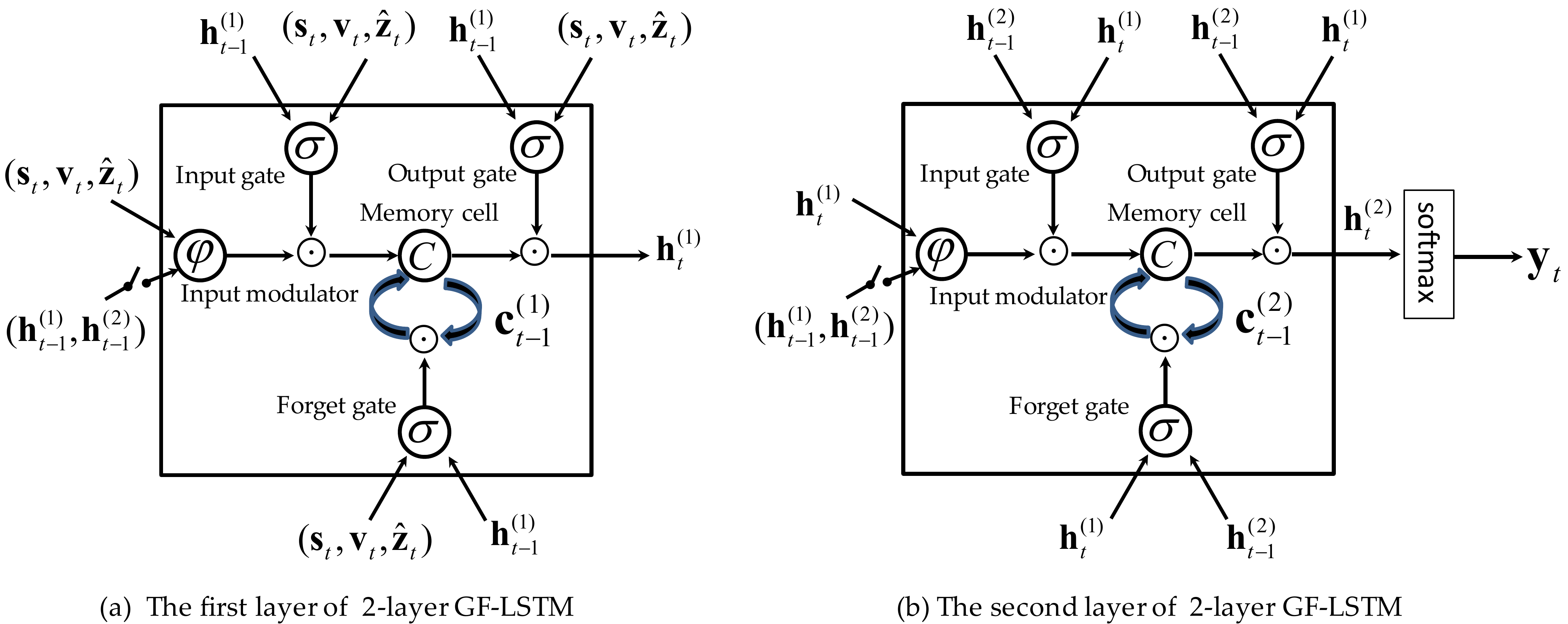}\\
  \caption{The diagram of the 2-layer GF-LSTM unit used in our 3G model. (a) denotes the first layer and (b) denotes the second layer.
   At the first layer, each gate has 4 input vectors: the global feature at the time-step $t$ ${\bold{v}}_t$, the local image feature at the time-step $t$ ${{\bf{\hat z}}_t}$, word representation at the time-step $t$ ${{\bf{s}}_t}$ and the hidden state of the first layer at time-step $t-1$ ${\bf{h}}_{t-1}^{(1)}$. When calculating the input modulation gate, not only the $({\bold{v}}_t,{{\bf{\hat z}}_t},{{\bf{s}}_t}$) is needed, but also the previous hidden states $({\bf{h}}_{t-1}^{(1)},{\bf{h}}_{t-1}^{(2)})$ is used.
   At the second layer, the previous hidden state of the second layer ${\bf{h}}_{t-1}^{(2)}$ and the current hidden state of the first layer ${\bf{h}}_{t}^{(1)}$ are as the input of the three gates. Both the previous states $({\bf{h}}_{t-1}^{(1)},{\bf{h}}_{t-1}^{(2)})$ need to be controlled by a gate defined in Eq. (\ref{gate-defining}) before they are imported into the input modulator. This is the core idea of the GF-LSTM.}
  \label{2_GF-LSTM}
\end{figure*}
\subsection{Gated Feedback LSTM for Sentence Generation }\label{GF-LSTM}

Two-layer gated feedback LSTM is used as the sentence generator in our method. In other words, RNN model in Fig. \ref{3G_fig} is two-layer gated feedback LSTM. Differing from Fig. \ref{GF-RNN}.(b), every LSTM unit has three inputs: $\bf{v}_t$, ${{\bf{\hat z}}_t}$ and $\bf{s}_t$.

\subsubsection{LSTM Model}
 Because the input of model is more complex than the GF-LSTM introduced in Section \ref{GF-RNNs} and GF-LSTM model is core in our model, in this subsection, we introduce our formulas in detail. Fig. \ref{2_GF-LSTM} shows the diagram of 2-layer GF-LSTM in our 3G model. The function $\varphi$ denotes the hyperbolic tangent function (\textsl{i.e.} $\varphi (x) = \tanh (x) = \frac{{{e^x} - {e^{ - x}}}}{{{e^x} + {e^{ - x}}}} = 2 \sigma(x) - 1$). First, according to the structure shown in Fig. \ref{2_GF-LSTM}, three gates are rewritten as follows:
 \begin{equation}\label{in}
\left\{ {\begin{array}{*{20}{l}}
{{\bf{i}}_t^{(1)} = \sigma \left( {W_i^{(1)}{{\bf{s}}_t} + U_i^{(1)}{\bf{h}}_{t - 1}^{(1)} + V_i^{(1)}{{\bf{v}}_t} + Z_i^{(1)}{{{\bf{\hat z}}}_t} + {\bf{b}}_i^{(1)}} \right)}\\
{{\bf{i}}_t^{(2)} = \sigma \left( {W_i^{(2)}{\bf{h}}_t^{(1)} + U_i^{(2)}{\bf{h}}_{t - 1}^{(2)} + {\bf{b}}_i^{(2)}} \right)}
\end{array}} \right.,
 \end{equation}

 \begin{equation}\label{forget}
\left\{ {\begin{array}{*{20}{l}}
{f_t^{(1)} = \sigma \left( {W_f^{(1)}{{\bf{s}}_t} + U_f^{(1)}{\bf{h}}_{t - 1}^{(1)} + V_f^{(1)}{{\bf{v}}_t} + Z_f^{(1)}{{{\bf{\hat z}}}_t} + {\bf{b}}_f^{(1)}} \right)}\\
{f_t^{(2)} = \sigma \left( {W_f^{(2)}{\bf{h}}_t^{(1)} + U_f^{(2)}{\bf{h}}_{t - 1}^{(2)} + {\bf{b}}_f^{(2)}} \right)}
\end{array}} \right.,
 \end{equation}

 \begin{equation}\label{out}
\left\{ {\begin{array}{*{20}{l}}
{{\bf{O}}_t^{(1)} = \sigma \left( {W_o^{(1)}{{\bf{s}}_t} + U_o^{(1)}{\bf{h}}_{t - 1}^{(1)} + V_o^{(1)}{{\bf{v}}_t} + Z_o^{(1)}{{{\bf{\hat z}}}_t} + {\bf{b}}_o^{(1)}} \right)}\\
{{\bf{O}}_t^{(2)} = \sigma \left( {W_o^{(2)}{\bf{h}}_t^{(1)} + U_o^{(2)}{\bf{h}}_{t - 1}^{(2)} + {\bf{b}}_o^{(2)}} \right)}
\end{array}} \right.,
 \end{equation}
where ${W_{*}^{(i)}}\in {\mathbb{R}^{h\times h}}$, ${U_{*}^{(i)}}\in {\mathbb{R}^{h\times h}}$, ${V_{*}^{(1)}}\in {\mathbb{R}^{h\times h}}$ and ${Z_{*}^{(1)}}\in {\mathbb{R}^{h\times h}}$ denote weights matrixes and ${\bf{b}}_{*}^{(i)}$ stands for biases. ${\bf{v}_t}\in {\mathbb{R}^{h}}$ and ${{\bf{\hat z}}_t}\in {\mathbb{R}^{h}}$ are the global and local image features, respectively. Their calculating formulas are introduced in the next two subsections.

\par Formulas of the current memory and hidden state are the same as Eqs. (\ref{cell})---(\ref{hidden}). The updating memory contents are computed as follows:
\begin{equation}\label{memory}
\left\{ {\begin{array}{*{20}{l}}
{{\bf{\tilde c}}_t^{(1)} = \tanh \left( {W_c^{(1)}{{\bf{s}}_t} + \sum\limits_{i = 1}^2 {g_h^{i \to 1}U_c^{i \to 1}{\bf{h}}_{t - 1}^{(i)} + V_c^{(1)}{{\bf{v}}_t} + Z_c^{(1)}{{{\bf{\hat z}}}_t} + {\bf{b}}_c^{(1)}} } \right)}\\
{{\bf{\tilde c}}_t^{(2)} = \tanh \left( {W_c^{(2)}{\bf{h}}_t^{(1)} + \sum\limits_{i = 1}^2 {g_h^{i \to 2}U_c^{i \to 2}{\bf{h}}_{t - 1}^{(i)}}  + {\bf{b}}_c^{(2)}} \right)}
\end{array}} \right.,
\end{equation}
where gate $g_h^{i \to j}$ is computed in Eq. (\ref{gate-defining}). ${W_c^{(i)}}\in {\mathbb{R}^{h\times h}}$, ${U_c^{i \to j}}\in {\mathbb{R}^{h\times h}}$, ${V_c^{(1)}}\in {\mathbb{R}^{h\times h}}$ and ${Z_c^{(1)}}\in {\mathbb{R}^{h\times h}}$ are weights matrixes. ${\bf{b}}_c^{(i)}\in {\mathbb{R}^{h}}$ are biases.

\par The LSTM module outputs a probability at each time-step. We write it as the following formulas:
\begin{equation}\label{lstm-output}
{{\bf{y}}_t} = {W_y}{\bf{h}}_t^{(2)} + {{\bf{b}}_y},
\end{equation}
\begin{equation}\label{lstm-p}
{{\bf{p}}_{t + 1}} = {\rm{softmax}}({{\bf{y}}_t}),
\end{equation}
where ${W_y}\in {\mathbb{R}^{N_0\times h}}$ and ${{\bf{b}}_y}\in {\mathbb{R}^{N_0}}$ denote passing forward parameters. ${{\bf{y}}_t}\in {\mathbb{R}^{N_0}}$ is an output of LSTM at the $t$-th time-step. ${{\bf{p}}_{t+1}}\in {\mathbb{R}^{N_0}}$ is a probability vector, of which each element represents the predicting probability of the corresponding word.

\par Having built the GF-LSTM model, initializing the system is another important thing to do. The memory and the hidden state are initialized by the following formulas:
\begin{equation}\label{init_m}
{{\bf{c}}_0} = \tanh \left( {{W_{c\_init}}\left( {\frac{1}{C}\sum\limits_{i = 1}^C {{{\bf{v}}_{li}}} } \right) + {{\bf{b}}_{c\_init}}} \right),
\end{equation}
\begin{equation}\label{init_h}
{{\bf{h}}_0} = \tanh \left( {{W_{h\_init}}\left( {\frac{1}{C}\sum\limits_{i = 1}^C {{{\bf{v}}_{li}}} } \right) + {{\bf{b}}_{h\_init}}} \right),
\end{equation}
where ${W_{c\_init}}\in {\mathbb{R}^{h\times h}}$ and ${W_{h\_init}}\in {\mathbb{R}^{h\times h}}$ are initial weights.  ${{\bf{b}}_{c\_init}}\in {\mathbb{R}^h}$ and ${{\bf{b}}_{h\_init}}\in {\mathbb{R}^h}$ are initial biases.
\subsubsection{Gate for Global Image Feature}
 \par ${{\bf{v}}_t}$ occurs several times in Section \ref{GF-LSTM}. And it is an output of global image feature controlled by gate. In previous works \cite{mao2014m-rnn, DBLP:journals/corr/KirosSZ14, vinyals2015nic, DBLP:journals/pami/VinyalsTBE17, karpathy2015devs, DBLP:journals/pami/KarpathyF17}, the vast majority of them import the global feature defined in Eq. (\ref{global feature}) at the first time-step or at each time-step into the RNN decoder, but Vinyals \textsl{et al.} \cite{vinyals2015nic, DBLP:journals/pami/VinyalsTBE17} find that global feature imported at the first time-step is better than that at every time-step. They explain that global feature imported at each time-step may bring more noise to the system. But images in the benchmark image-caption datasets are high quality with little noise. Therefore, this reason may be a little far-fetched. In this paper, we want to design a robust algorithm that can autonomously decide when and how much the global feature should be imported into the decoder. Inspired by the gating mechanism exploited in LSTM, we design a gate before the global feature is imported into the multi-modal embedding part. The gate is defined as follows:
 \begin{equation}\label{gate-global}
 {g_t} = \sigma ({\bf{w}}_g^T{\bf{h}}_{t - 1}^{(2)} + {b_g}),
 \end{equation}
 where ${{\bf{w}}_g}\in {\mathbb{R}^h}$ is the weight vector, $b_g$ is the bias. So the $t$-th gate $g_t$ is a scaler and its value correlates with the previous time-step 2-layer hidden state ${\bf{h}}_{t - 1}^{(2)}$.
 \par After calculating the gate, the global image feature at time-step $t$ is computed as follows:
 \begin{equation}\label{gg}
 {{\bf{v}}_t} = {g_t}{\bf{v}}.
 \end{equation}
 \par Through Eq. (\ref{gg}), if we set $g_t = 1$ at $t = 0,1,\cdots,T$, $\bf{v}$ is imported into the decoder at each time-step. If we set $g_t = 1$ at $t = 0$ and $g_t = 0$ at $t = 1,2,\cdots,T$, $\bf{v}$ is only imported into the decoder at the first time-step.
 Theoretically speaking, the 3G model proposed in this paper is more general and more robust.
\subsubsection{Attention Mechanism for Local Image Feature}
\par The local image feature ${\bf{\hat z}}_t$ denotes the local information of image. Here we use the attention mechanism as introduced in \cite{DBLP:conf/icml/XuBKCCSZB15} for local feature. At each time-step, the attention mechanism uses the previous hidden state $h_{t - 1}^{(2)}$ to decide the local feature. The attention model is defined as follows:
\begin{equation}\label{att}
{{\bf{\tilde \alpha }}_t} = \tanh \left[ {{{\left( {{\bf{w}}_a^T{{\bf{v}}_l}} \right)}^T} + U_a{\bf{h}}_{t - 1}^{(2)} + {{\bf{b}}_a}} \right],
\end{equation}
\begin{equation}\label{att-p}
{{\bf{\alpha }}_t} = {\rm{softmax}} \left( {{{{\bf{\tilde \alpha }}}_t}} \right),
\end{equation}
where ${{\bf{w}}_a}\in {\mathbb{R}^{h}}$ and ${U_a}\in {\mathbb{R}^{h\times h}}$ are weights. ${\bf{v}}_l$ is defined in Eq. (\ref{local feature}). ${{\bf{b}}_a}\in {\mathbb{R}^{h}}$ is the bias. The annotation vector ${{\bf{\alpha }}_t} \buildrel \Delta \over = {\left[ {\begin{array}{*{20}{c}}{{\alpha _{t1}}}& \cdots &{{\alpha _{tC}}}\end{array}} \right]^T} \in {\mathbb{R}^{C}}$ is a probability vector whose each dimension value denotes the probability of the corresponding local image feature. In our algorithm, we use the soft attention model. Therefore, ${{\bf{\hat z}}_t}$ is calculated as follows:
\begin{equation}\label{lv}
{{\bf{\hat z}}_t} = {{\bf{v}}_l}{{\bf{\alpha }}_t} = \sum\limits_{i = 1}^C {{\alpha _{ti}}{{\bf{v}}_{li}}} .
\end{equation}
\par Through Eq. (\ref{lv}), the regions are selected at time-step $t$ by the annotation vector ${{\bf{\alpha }}_t}$.
\par After designing the model and combining the objective function in Eq. (\ref{OF}), the loss function of our model can be written as follows:
\begin{equation}\label{LF}
L(\theta ) =  - \log P\left( {S\left| I \right.} \right),
\end{equation}
where ${\theta}$ is the parameter set which includes parameters of the GF-LSTM, embedding matrixes in Section \ref{Image Representation} and Section \ref{Sentence Representation} and all gate models.
\par The proposed model is trained with \textsl{back-propagation through time} (BPTT) algorithm \cite{fairbank2013equivalence, werbos1990backpropagation} to minimize the cost function $L(\theta)$.

\section{Experiments} \label{Experiments}

In this section, we begin by describing the publicly available datasets used for training and testing the model and the evaluating metrics for image caption generation. Then some typical state-of-the-art models are simply introduced. Finally, we show the quantitative results compared with the recent state-of-the-art methods and analyze the experimental results.
\subsection{Datasets}\label{Dataset}
In this subsection, three benchmark datasets are introduced. They are Flickr8K \cite{rashtchian2010flickr8k}, Flickr30K \cite{young2014flickr30k} and MS COCO \cite{lin2014coco}. Among them, Flickr8K and Flickr30K have 8,092 and 31,783 images respectively, and each image has 5 reference sentences. The images in these two datasets focus on people and animals performing some actions. The most challenging MS COCO datast has 82,783 images and most of the images have 5 reference sentences, but there are also some images have references in excess of 5.

\subsection{Evaluation Metrics}
In order to evaluate the proposed method, two objective metrics are used in this paper.
They are BLEU \cite{papineni2002bleu} and METEOR \cite{banerjee2005meteor}. These two metrics are originally designed for evaluating the quality of the automatically machine translation. BLEU score represents the precision ratio of the generated sentence compared with the reference sentences. METEOR score reflects the precision and recall ratio of the generated sentence. It is based on the harmonic mean of uniform precision and recall. For BLEU, we use the scores from BLEU-1 to BLEU-4, which denote the precision of N-gram (N equals to 1, 2, 3 and 4). For both two metrics, the higher score they are, the higher quality of the generated sentences they have.

\subsection{Comparison Models}\label{comparison model}
In this subsection, some typical state-of-the-art models are briefly introduced. These  models are both using \emph{CNN + RNN} diagram which is the most effective diagram for image caption generation, but they have some differences in detail.
\begin{itemize}
  \item \textbf{m-RNN \cite{mao2014m-rnn}.} \textsl{Multi-modal recurrent neural network} (m-RNN) contains three parts: a vision part, a language module part and a multi-modal part. The vision part is a pre-trained deep CNN to extract the feature of the images. The language model encodes each word in the dictionary and stores the semantic temporal context. The multi-modal part connects the image representation and word embedding together by a  one-layer representation. This model imports the image representation into the multi-modal module at each time-step. The structure of LRCN \cite{donahue2015lrcn} is similar to this model except the language model: LRCN uses LSTM as the language model while m-RNN uses the ``vanilla'' RNN.
  \item \textbf{Google-NIC \cite{vinyals2015nic, DBLP:journals/pami/VinyalsTBE17}.} Unlike m-RNN, Google-NIC just projects the image feature into the embedding space and imports it into the RNN at the first time-step. Therefore, the RNN here is not only the language model, but also the multi-modal model. DeVS \cite{karpathy2015devs, DBLP:journals/pami/KarpathyF17} is very similar to this model, but they also have a little difference, where LSTM is used as sentence generator in Google-NIC but ``vanilla'' RNN is used in DeVS.
  \item \textbf{NIC-VA \cite{DBLP:conf/icml/XuBKCCSZB15}.} This model has several variations in different tasks such as image caption generation, machine translation, video clip description and speech recognition and has achieved great success in these tasks. In image caption generation task, the attention model uses the output of the CNN convolutional layer as the image representation. Through flatten operation, every vector stands for one local image feature. These features would be selected by the attention mechanism and input into RNN at each time-step. This model properly draws on the human attention mechanism, so it gets a great success.
\end{itemize}

\subsection{Experiment Setup} \label{setup}
\subsubsection{Dataset Processing}
 Before the experiment, we have preprocessed the datasets as \cite{karpathy2015devs} did. At first, we convert all letters of sentences to lowercase and remove non-alphanumeric characters. Then we get rid of words that occur less than five times on the training set. Because some images in MS COCO have more than 5 corresponding sentences, we discard these data to grantee every image has the same number of describing sentences. Since  the ground-truth captions of the MS COCO test set are blind to the public, we use the publicly released splits\footnotemark \footnotetext{\url{https://cs.standford.edu/people/karpathy/deepimagesent}} which is used in DeVS within 5,000 testing images.
 \subsubsection{Image Feature}
 In the proposed model, deep features generated from the CONV5-4 and FC7 layers of VGG-19 are used to represent the images. In our experiments, the global image feature is generated from the FC7 layer. Therefore, the global feature is a  4096-dimension vector (\textsl{i.e.} $l$ in Section \ref{gv} equals to 4096). The local image feature output from the CONV5-4 layer.  Through flattening, the local feature set of one image has 196 vectors with a dimension of 512 (\textsl{i.e.} $C=196$, $D=512$ in Section \ref{llv}).
 \subsubsection{Word Encoding}
 In our model, we encode words into one-hot vectors. For example, the benchmark dataset has $M$ different words, every word is encoded into a $M$-dimension vector, in which only one value equals to 1 and others equal to 0. So the location of 1 in the vector denotes the corresponding word in the dictionary. It implies that $N_{0}$ in Section \ref{Sentence Representation} equals to $M$.
 \subsubsection{Training Option}
 The proposed model is trained with \textsl{stochastic gradient descent} (SGD) using adaptive learning rate algorithms. Similar to \cite{DBLP:conf/icml/XuBKCCSZB15}, the RMSProp algorithm is used for the Flickr8K dataset and for Flickr30K/MS COCO.

\subsection{Results Evaluation and Analysis} \label{eperimental results}
\begin{table*}[ht] \normalsize \addtolength{\tabcolsep}{-3.5pt}\setlength{\tabcolsep}{0.6pt}
\setlength{\abovecaptionskip}{10pt}
\setlength{\belowcaptionskip}{5pt}
\begin{spacing}{1.5}
\newcommand{\tabincell}[2]{\begin{tabular}{@{}#1@{}}#2\end{tabular}}
\centering
\caption{Performance of different models on Flickr8K, Flickr30K \& MSCOCO.}
\label{Generating Results}
\resizebox{\linewidth}{!}{
\begin{tabular}{|l|ccccc|ccccc|ccccc|}
  \hline
 &\multicolumn{5}{c|}{Flickr8K}  &\multicolumn{5}{c|}{Flickr30K} &\multicolumn{5}{c|}{MS COCO}\\
  \cline{2-16}\raisebox{0.5em}{Model} & \tabincell{c}{B-1} & \tabincell{c}{B-2} & \tabincell{c}{B-3} & \tabincell{c}{B-4} & \tabincell{c}{M} & \tabincell{c}{B-1} & \tabincell{c}{B-2} & \tabincell{c}{B-3} & \tabincell{c}{B-4}  & \tabincell{c}{M} & \tabincell{c}{B-1} & \tabincell{c}{B-2} & \tabincell{c}{B-3} & \tabincell{c}{B-4}  & \tabincell{c}{M}\\
  \hline
  \hline
  \multicolumn{16}{|l|}{global image feature based}\\
  \hline
  \raisebox{0em} {m-RNN \cite{mao2014m-rnn}} & 56.5 & 38.6 & 25.6 & 17 & - & 60 & 41 & 28 & 19 & - & 66.8 & 48.8 & 34.2 & 23.9 & 22.1\\
  \raisebox{0em} {DeVS \cite{DBLP:journals/pami/KarpathyF17}} & 57.9 & 38.3 & 14.5 & 16 & 16.7 & 57.3 & 36.9 & 24 & 15.7 & 15.3 & 62.5 & 45 & 32.1 & 23 & 19.5\\
  \raisebox{0em} {LRVR} \cite{chen2015lrvr} & - & - & - & 14.1 & 18 & - & - & - & 12.6 & 16.4 & - & - & - & 19 & 20.4\\
  \raisebox{0em} {Google-NIC \cite{DBLP:journals/pami/VinyalsTBE17}} & 63 & 41 & 27 & - & - & 66.3 & 42.3 & 27.7 & 18.3 & - & 66.6 & 46.1 & 32.9 & 24.6 & 23.7\\
  \raisebox{0em} {LRCN \cite{7558228}} & - & - & - & - & - & 58.8 & 39.1 & 25.1 & 16.5 & - & 62.8 & 44.2 & 30.4 & 21 & -\\
  \hline
  \hline
  \multicolumn{16}{|l|}{attention-based}\\
  \hline
  \raisebox{0em} {NIC-VA \cite{DBLP:conf/icml/XuBKCCSZB15}} & 67 & 44.8 & 29.9 & 19.5 & 18.9 & 66.7 & 43.4 & 28.8 & 19.1 & 18.4 & 68.9 & 49.2 & 34.4 & 24.3 & 23.9\\
  \raisebox{0em} {ATT-FCN \cite{DBLP:conf/cvpr/YouJWFL16}} & - & - & - & - & - & 64.7 & 46.0 & 32.4 & 23.0 & 18.9 & 70.9 & 53.7 & 40.2 & 30.4 & 24.3\\
  \raisebox{0em} {(RA+SS)-ENSEMBLE \cite{7792748}} & - & - & - & - & - & 64.9 & \textbf{46.2} & 32.4 & 22.4 & 19.4 & \textbf{72.4} & \textbf{55.5} & \textbf{41.8} & \textbf{31.3} & 24.8\\
  \hline
  \hline
  \raisebox{0em} {3G} & \textbf{69.9} & \textbf{48.5} & \textbf{34.4} & \textbf{23.5} & \textbf{22.3} & \textbf{69.4} & 45.7 & \textbf{33.2} & \textbf{22.6} & \textbf{23.0} & 71.9 & 52.9 & 38.7 & 28.4 & \textbf{24.3}\\
  \hline

\end{tabular}}
\end{spacing}
\end{table*}

Table \ref{Generating Results} gives a summary performance of different models on the three benchmark datasets. Capital letters M in Table \ref{Generating Results} stands for the METEOR score. Setting on MS COCO has some minor differences among the compared models, because the test set of MS COCO has no given reference sentences. There is no standard split. For example, DeVS \cite{karpathy2015devs, DBLP:journals/pami/KarpathyF17}, LRCN \cite{donahue2015lrcn} and NIC-VA \cite{DBLP:conf/icml/XuBKCCSZB15} isolate 5,000 images from the validation set as testing set, m-RNN chooses 4,000 validation images and 1,000 testing data from the validation set, Google-NIC \cite{vinyals2015nic, DBLP:journals/pami/VinyalsTBE17} selects 4,000 images from the validation sets as testing set while LRVR \cite{chen2015lrvr} tests its model on the validation set with 1,000 images. In fact, the more testing data, the more challenging for the proposed models are. 5000 testing images are used in our experiment.  The experimental results demonstrate that our 3G model is better than other state-of-the-art models. The results of the other compared models excepting NIC-VA are transcribed from their corresponding articles and we reproduce the results of the NIC-VA with the released code\footnotemark \footnotetext{\url{https://github.com/kelvinxu/arctic-captions}}.

\par Through the experimental results in Table \ref{Generating Results}, it can be observed that our 3G model almost gets the highest score on Flickr8K and Flickr30K. On the MS COCO dataset, the (RA+SS)-ENSEMBLE \cite{7792748} gets a higher score on BLEU-2 to BLEU-4 than ours. But our 3G model shows a better performance on BLEU-1 and METEOR. Moreover, (RA+SS)-ENSEMBLE \cite{7792748} needs to obtain a scene vector for each image. More specially, it needs to use \textsl{Latent Dirichlet Allocation} (LDA) to obtain a ``scene vector" for each image. The model also need to train a multilayer perceptron to predict the scene vector. Therefore, (RA+SS)-ENSEMBLE is more complex than the proposed model in this paper.
Table \ref{Generating Results} shows that the attention-based methods are more effective than the global image feature based methods. This means the attention mechanism is more effective than other models only using the global image feature.
However, the 3G model proposed in this paper shows a better or at least comparable performance, which confirms the effectiveness of the proposed model. The quantitative results in Table \ref{Generating Results} proves that the global image feature using gating mechanism and the gated feedback LSTM for multi-modal embedding plays an important role on the image caption generation task. Three main points make our 3G  model better than the other state-of-the-art models. First, the proposed 3G model in this paper exploits the visual attention mechanism. Second, the introduction of the global image feature with gating control is a good supplement because the visual attention part mainly focuses on the local of the given image. Third, the 2-layer GF-LSTM makes our language model stronger than other language models in the contrast models.
\begin{table*}[ht]
\setlength{\abovecaptionskip}{10pt}
\setlength{\belowcaptionskip}{5pt}
\begin{spacing}{1.5}
\newcommand{\tabincell}[2]{\begin{tabular}{@{}#1@{}}#2\end{tabular}}
\centering
\caption{Automatic metric scores on the MSCOCO test server.}
\label{COCO_sever}
\resizebox{\linewidth}{!}{
\begin{tabular}{lcccccccccc}
  \hline
 &\multicolumn{5}{l}{5-Refs}  &\multicolumn{5}{l}{40-Refs} \\
  \cline{2-11}\raisebox{0.5em}{Model} & \tabincell{c}{B-1} & \tabincell{c}{B-2} & \tabincell{c}{B-3} & \tabincell{c}{B-4} & \tabincell{c}{M} & \tabincell{c}{B-1} & \tabincell{c}{B-2} & \tabincell{c}{B-3} & \tabincell{c}{B-4}  & \tabincell{c}{M} \\
  \hline
  \hline
 \raisebox{0em} {Human} & 66.3 & 46.9 & 32.1 & 21.7 & 25.2 & 88.0 & 74.4 & 60.3 & 47.1 & 33.5 \\
  \raisebox{0em} {DeVS \cite{DBLP:journals/pami/KarpathyF17}} & 65.0 & 46.4 & 32.1 & 22.4 & 21.0 & 82.8 & 70.1 & 56.6 & 44.6 & 28.0 \\
  \raisebox{0em} {MSR \cite{DBLP:conf/cvpr/FangGISDDGHMPZZ15}} & 69.5 & 52.6 & 39.1 & 29.1 & 24.7 & 88.0 & 78.9 & 67.8 & 56.7 & 33.1 \\
  \raisebox{0em} {NN \cite{DBLP:journals/corr/DevlinGGMZ15}} & 70 & 52 & 38 & 28 & 24 & 87 & 77 & 66 & 54 & 32 \\
  \raisebox{0em} {MLBL \cite{DBLP:conf/icml/KirosSZ14}} & 67 & 50 & 36 & 26 & 22 & 85 & 75 & 63 & 52 & 33.1 \\
  \raisebox{0em} {NIC-VA \cite{DBLP:conf/icml/XuBKCCSZB15}} & 70.5 & 52.8 & 38.3 & 27.7 & 24.1 & 88.1 & 77.9 & 65.8 & 53.7 & 32.2 \\
  \hline
 \raisebox{0em} {RA \cite{7792748}} & 72.2 & 55.6 & 41.8 & 31.4 & 24.8 & 90.2 & 81.7 & 71.1 & 60.1 & 33.6 \\
 \raisebox{0em} {ATT \cite{DBLP:conf/cvpr/YouJWFL16}} & 73.1 & 56.5 & 42.4 & 31.6 & 25.0 & 90 & 81.5 & 70.9 & 59.9 & 33.5 \\
 \raisebox{0em} {Att \cite{7934440}} & 73 & 56 & 41 & 31 & 25 & 89 & 80 & 69 & 58 & 33 \\
 \hline
  \raisebox{0em} {3G} & 70.1 & 53.2 & 39.3 & 29.1 & 23.4 & 88.2 & 79.0 & 68 & 56.9 & 31.7 \\
  \hline

\end{tabular}}
\end{spacing}
\end{table*}

\par MS COCO team hosts a test server allowing people to evaluate their models online \footnotemark \footnotetext{\url{https://competitions.codalab.org/competitions/3221}}. The evaluation is on the test set, of which the reference sentences are blind to the public. Each image in the test set is labelled with 40 sentences. We evaluate the proposed model on the test server and Table \ref{COCO_sever} shows the performance of the published state-of-the-art image captioning models on the online COCO test server. The test results are split into two categories: 5-Refs and 40-Refs, which respectively denote the results for 5 reference sentences and 40 reference sentences.
 In Table \ref{COCO_sever}, the compared methods can be divided into two categories. 1) ``CNN-RNN'' diagram and 2) Detector+``CNN-RNN'' diagram. In the first category, only image feature is used to generate the sentence. However, the second category methods need extra image information attained by the extra detector. Therefore, DevS \cite{DBLP:journals/pami/KarpathyF17}, MSR \cite{DBLP:conf/cvpr/FangGISDDGHMPZZ15}, MLBL \cite{DBLP:conf/icml/KirosSZ14}, NIC-VA \cite{DBLP:conf/icml/XuBKCCSZB15} and our 3G belong to the first category; while RA \cite{7792748}, ATT \cite{DBLP:conf/cvpr/YouJWFL16} and Att \cite{7934440} belong to the second category. Through analyzing the results in Table \ref{COCO_sever}, we can draw two conclusions. First, the proposed method in this paper shows the best performance among the first kind of methods. In other words, our method shows the best performance when only the image feature is used. Specially, compared with the spatial attention method, NIC-VA, almost all the scores are improved by our method (except the B-1 for 5-Refs and METEOR). Second, the methods under the Detector+``CNN-RNN'' diagram are obviously better than the first kind of methods. Because they utilize more image information to generate the describing sentence. For instance, object detector is used before the image feature representation in RA model. The detector needs be pre-trained on the extra object detecting dataset. For ATT and Att, extra image attribute information needs to be input into the sentence generator and the attribute detector is trained on the pre-labeled dataset. Therefore, RA, ATT and Att reasonably outperform the proposed method. Meanwhile, it also implies that our method still has potential for improvement if more image information is used.

\begin{table}[!htb] \normalsize \addtolength{\tabcolsep}{-3.5pt}\setlength{\tabcolsep}{5.5pt}
\setlength{\abovecaptionskip}{10pt}
\setlength{\belowcaptionskip}{5pt}
\begin{spacing}{1.5}
\newcommand{\tabincell}[2]{\begin{tabular}{@{}#1@{}}#2\end{tabular}}
\centering
\caption{Experimental results on MS COCO to verify the effectiveness different deep image features.}
\label{Dif_feature}
\small\begin{tabular}{lccccc}
  \hline
  \raisebox{0em} {Model} & B-1 & B-2 & B-3 & B-4 & METEOR\\
  \hline
  \hline
  \raisebox{0em} {3G (AlexNet)} & 63.2 & 48.1 & 33.2 & 22.5 & 18.3\\
  \raisebox{0em} {3G (GoogleNet)} & 70.8 & \textbf{52.3} & \textbf{39.1} & 28.2 & 22.9\\
  \raisebox{0em} {3G (VGG)} & \textbf{71.6} & 52.2 & 39.0 & \textbf{28.9} & \textbf{23.8}\\
  \hline

\end{tabular}
\end{spacing}
\end{table}

In the experiments, we choose the VGG-Net as the image feature extractor. The main reasons are as follows: 1) almost all the state-of-the-art contrast methods compared in the experiments use the VGG-Net as image feature extractor \cite{karpathy2015devs, mao2014m-rnn, kiros2014mnlm, DBLP:conf/icml/XuBKCCSZB15}; 2) VGG-Net is considered as one of the most effective feature extractor which can extract the discriminative and expressive feature for image. However, we also add an extra experiment to verify the effectiveness of the different image features and the results are shown in Table \ref{Dif_feature}. Among the models in Table \ref{Dif_feature}, the image features are extracted by the corresponding network in the parentheses. For AlexNet and VGG-Net, the feature maps output by the last convolutional layer are used as the local image features and the feature vector output by the FC-7 layer is used as the global image feature. For GoogleNet, we use the output from the Inception (4c) as the local feature and the AVG-POOL layer's output as the global image feature. As can be seen in Table \ref{Dif_feature}, the model with AlexNet is less powerful than the model with GoogLeNet or VGG-Net. This is mainly because GoogLeNet and VGG-Net can learn the discriminative and representative features in a hierarchical manner. Model with VGG-Net gets higher scores on B-1, B-4 and METEOR metrics than the model with GoogLeNet.

\begin{figure*}[ht]
  \centering
  \includegraphics[width=0.95\linewidth]{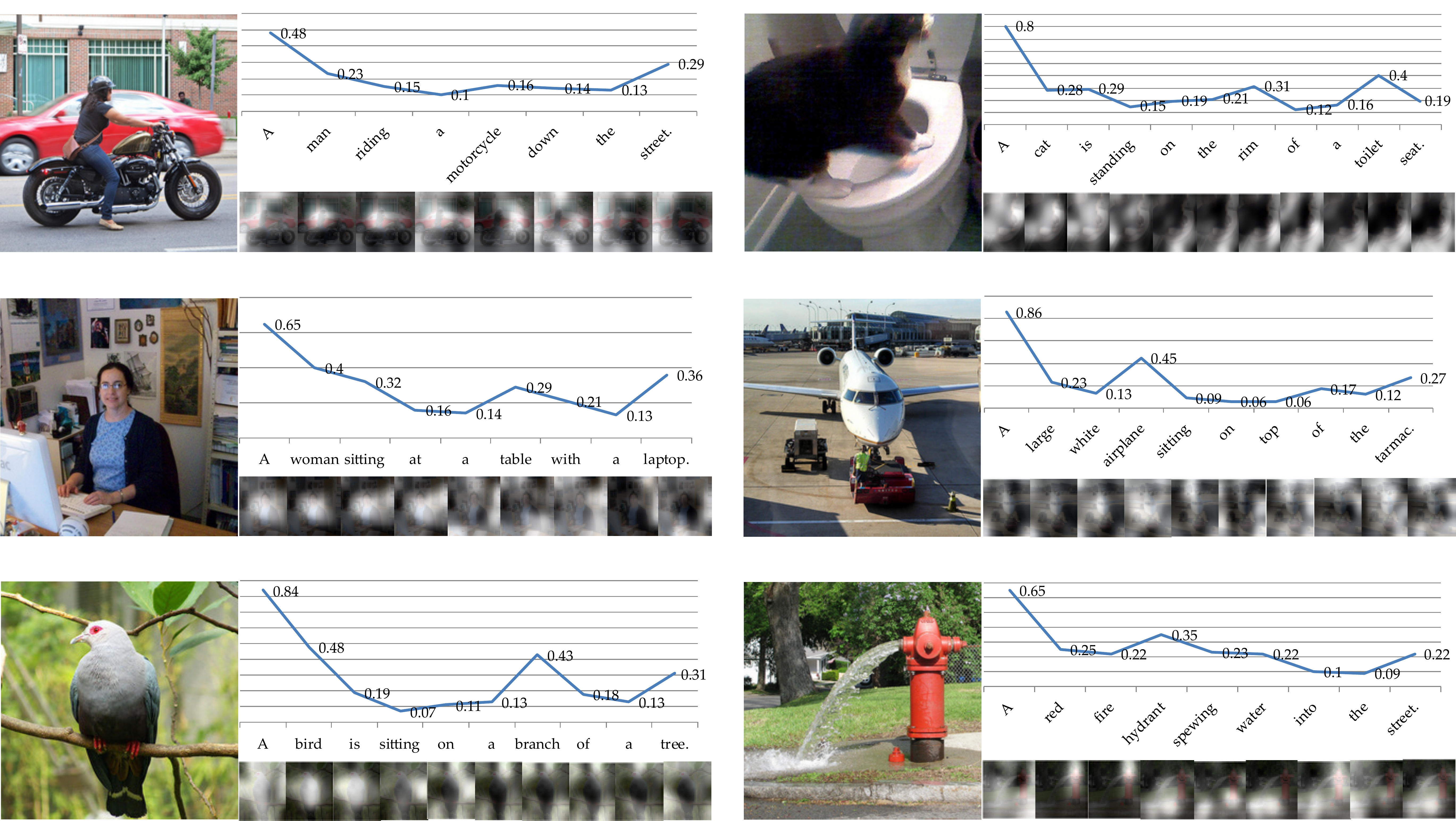}\\
  \caption{Visualization of attending to the correct object on the MS COCO dataset. White indicated the attended regions which are corresponded to the words. The line chart denotes the values of the global image feature gates corresponding to the words.}
  \label{visualization}
\end{figure*}

Fig. \ref{visualization} shows the generated captions with the spatial attention maps and the value of the global image feature gates. We can see that the first word and the object words have relatively larger values and the non-visual words have relatively smaller values. It confirms that the non-visual words (except the first word) such as ``a'' and ``on'' need less visual information (with smaller values of global image feature gates). Conversely, when generating a visual word such as ``airplane'', it needs more visual information (with a larger value of global image feature gate).

\begin{figure*}[ht]
  \centering
  \includegraphics[width=0.95\linewidth]{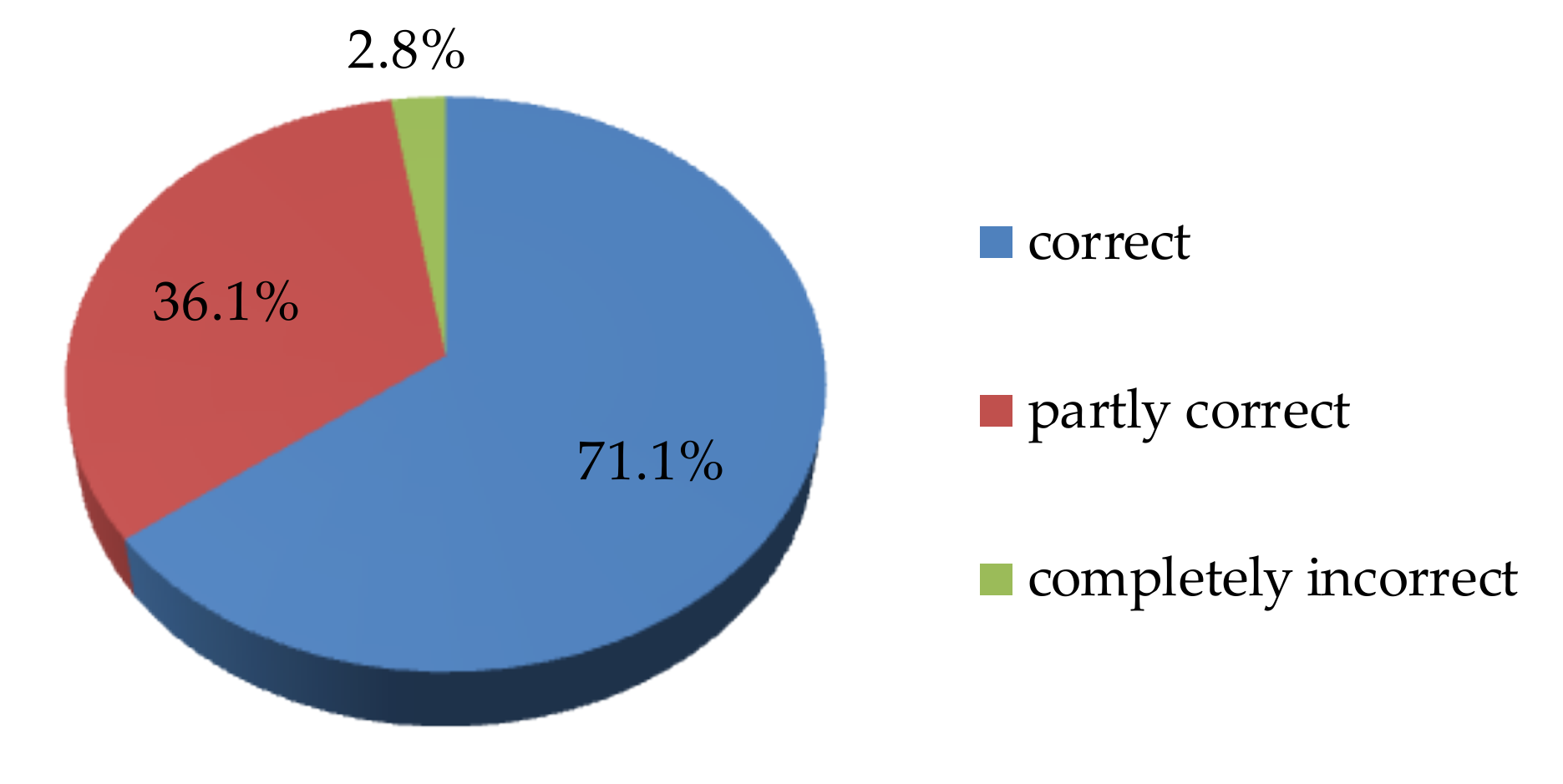}\\
  \caption{Human-based evaluation on the MS COCO test set. The quality of the generated sentences is divided into three categories: the sentence describes the image 1) correctly, 2) partly correct, or 3) completely incorrect.}
  \label{human evaluation}
\end{figure*}

We also add a crowdsourcing experiment on the MS COCO test set. To complete the human-based evaluation, we randomly sample 1000 images from the MS COCO test set. The corresponding sentences are generated by our trained model. After that, the image-caption pairs are evaluated by 5 persons. The quality of the sentences is divided into three categories: the sentence describes the image 1) correctly, 2) partly correct, or 3) completely incorrect.
When the sentence can describe all the content of the corresponding image, and it is also with spelling and grammatical correctness, this sentence should be judged to be correct. When the sentence can partly describe the content of the image, or it has small defects in spelling or grammar, this sentence should be judged partly correct. When the sentence cannot describe the content of the image, or it is unreadable, we consider this sentence is completely incorrect.
Every image-caption pair is marked with one of the categories by each person. At last, we compute the percentage of each category. The result is shown in Fig. \ref{human evaluation}. The figure shows that $71.1\%$ descriptions completely reflect the contents of the corresponding images. Only $2.8\%$ sentences are completely irrelevant to the images. Therefore, the human-based evaluation further validates the effectiveness of the proposed 3G model.

\par Furthermore, we want to know how much the global image feature using gating mechanism and the gated feedback LSTM for multi-modal embedding affect the model. Someone may also think the increase of the performance may be caused by the gated feedback LSTM but not the global image feature with gating control, because the gated feedback RNN is proved effective in language tasks \cite{DBLP:conf/icml/ChungGCB15}. To evaluate the effectiveness of both the gate for global image feature and the gated feedback LSTM, two groups of experiments have been done on MS COCO.

\begin{figure*}[ht]
  \centering
  \includegraphics[width=0.95\linewidth]{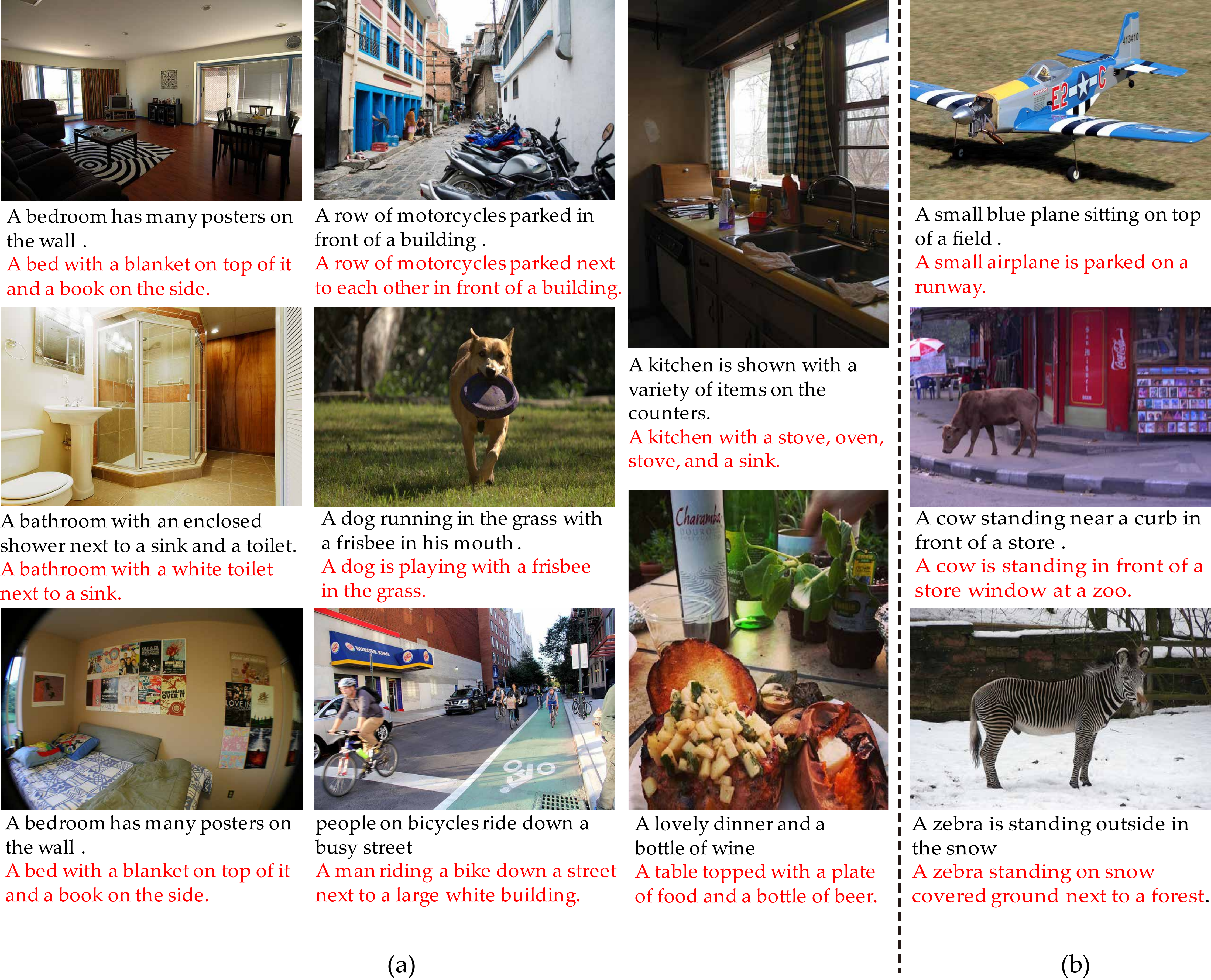}\\
  \caption{Examples of images and caption pairs from MS COCO dataset. The captions with \textcolor[rgb]{0.00,0.00,0.00}{black} font come from the reference sentences in the validation set and \textcolor[rgb]{1.00,0.00,0.00}{red} font are generated by our 3G model, respectively. (a) (on the left of the black dotted line) denotes the captions generated by our 3G model are proper with the given images. And on the contrary, (b) (on the right of the black dotted line) denotes the captions generated by our 3G model are not accurate enough to describe the content of the given images.}
  \label{examples}
\end{figure*}
\subsubsection{The effectiveness of image feature fusion with gated mechanism}
\begin{table}[!htb] \normalsize \addtolength{\tabcolsep}{-3.5pt}\setlength{\tabcolsep}{5.5pt}
\setlength{\abovecaptionskip}{10pt}
\setlength{\belowcaptionskip}{5pt}
\begin{spacing}{1.5}
\newcommand{\tabincell}[2]{\begin{tabular}{@{}#1@{}}#2\end{tabular}}
\centering
\caption{Experimental results on MS COCO to verify the effect of the feature fusion global and local image feature with gating control}
\label{Results 1}
\small\begin{tabular}{|l|ccccc|}
  \hline
  \raisebox{0em} {Model} & B-1 & B-2 & B-3 & B-4 & METEOR\\
  \hline
  \raisebox{0em} {Google-NIC} & 66.6 & 46.1 & 32.9 & 24.6 & 23.7\\
  \raisebox{0em} {LRCN} & 62.8 & 44.2 & 30.4 & 21 & -\\
  \raisebox{0em} {NIC-VA} & 68.9 & 49.2 & 34.4 & 24.3 & 23.9\\
  \hline
  \raisebox{0em} {GL-NIC} & \textbf{70.9} & \textbf{50.9} & \textbf{36.7} & \textbf{26.0} & \textbf{25.3}\\
  \hline

\end{tabular}
\end{spacing}
\end{table}
In order to verify the effect of the introduction of the image global feature with gating control, we use the 1 layer LSTM as the decoder, which we name it GL-NIC in Table \ref{Results 1}. Experimental results in Table \ref{Results 1} prove that the fusion of the global image feature and the local image feature is useful for image caption generation. In fact, the contrast models can be regarded as the special situations of GL-NIC : 1) when setting ${\alpha _t}=0$, ${g_t}=1$ at every time-step, GL-NIC degenerates as LRCN; 2) when setting ${g_t}=0$ for all $t$, GL-NIC degenerates as NIC-VA; 3) when setting ${\alpha _t}=0$ for all $t$, ${g_t}=1$ at $t=0$ and ${g_t}=1$ for other $t$, GL-NIC degenerates as Google-NIC. Therefore, fusing the global and local image features with gated mechanism can be more comprehensive and robust to describe the content of the image.

\subsubsection{The effectiveness of the gated feedback LSTM}
\begin{table}[!htb] \normalsize \addtolength{\tabcolsep}{-3.5pt}\setlength{\tabcolsep}{3.5pt}
\setlength{\abovecaptionskip}{10pt}
\setlength{\belowcaptionskip}{5pt}
\begin{spacing}{1.5}
\newcommand{\tabincell}[2]{\begin{tabular}{@{}#1@{}}#2\end{tabular}}
\centering
\caption{Experimental results  on MS COCO to verify the effect of the gated feedback LSTM}
\label{Results 2}
\small\begin{tabular}{|l|ccccc|}
  \hline
  \raisebox{0em} {Model} & B-1 & B-2 & B-3 & B-4 & METEOR\\
  \hline
  \raisebox{0em} {NIC-VA (2-layer LSTM)} & 70.1 & 50.3 & 35.7 & 25.5 & 24.6\\
  \hline
  \raisebox{0em} {NIC-VA (GF-LSTM)} & \textbf{71.6} & \textbf{51.5} & \textbf{37.2} & \textbf{26.5} & \textbf{25.5}\\
  \hline

\end{tabular}
\end{spacing}
\end{table}

To test and verify the effect of the gated feedback LSTM on MS COCO, the branch of the global image feature is discarded from the proposed 3G model. Then the 3G model is degenerated as NIC-VA with GF-LSTM sentence generator. So the model is named as NIC-VA (GF-LSTM) in Table \ref{Results 2}. We compared NIC-VA (GF-LSTM) model with NIC-VA model. However, the decoder in NIC-VA is 1-layer LSTM, for the sake of fairness, we changed the decoder in NIC-VA into 2-layer stacked LSTM. We mark this model as NIC-VA-2-LSTM. Table \ref{Results 2} shows that the performance of NIC-VA (GF-LSTM) is better than NIC-VA (2-layer LSTM). This is because the GF-LSTM can deal with the issue of learning multiple adaptive timescales. The gated feedback collecting method which is a strategy to increase the depth of the LSTM not only uses the previous equal level hidden state, but also uses the previous higher level hidden state. In fact, J. Chung \textsl{et al.} \cite{DBLP:conf/icml/ChungGCB15} has proven that the GF-RNN outperforms the traditional stacked RNN, especially as the number of nesting levels grows or the length of target sequences increases.

Fig. \ref{examples} shows some examples of image caption generation on the validation set of the MS COCO. The red font sentences are generated by our 3G model, while the black font sentences are given references which are annotated by human beings. According to Fig. \ref{examples} (a), our 3G model can accomplish the image caption generation task very well. Specially, our 3G model can not only generate the proper sentences to describe the main contents of the given images, but also give more elaborate descriptions for the given images. For example, the first image of the second column, the reference describes the relationship of the motorcycles and building (``motorcycles parked in front of a building''), while the sentence generated by our 3G model not only describe the relationship between them, but also describe the relationships between the motorcycles (``next to each other''). Another interesting but not intentional discovery is that our 3G model can generate grammatical correcting sentences, which may be ignored by humans. For instance, the third image of the second column, the first letter ``p'' should be capitalized but the reference gives the lower case. Some little errors like this will not be occur in our 3G model.
\par Some negative examples are also shown in Fig. \ref{examples} (b). Such as the third image in column 4, the generated sentence is ``A zebra standing on snow covered ground next to a forest'', but in fact only a few trees which cannot determine whether it is a forest. Despite that our model makes a small mistake on the scene background, our model always tries to describe all the content of one image. In fact, the sentence generated by 3G model describes the main contents of the image (``A zebra standing on snow covered ground''), but the background gets a little mistake (the zebra may not stand ``next to a forest''). The descriptions generated for the  first and the second images in column 4 also have some mistakes. We think this may be caused by the repeated scene and reference sentences. That is to say, when a plane in an image, the plane often accompanied by a runway. So the model has learned much knowledge like this, when importing an image similar but having some differences with this scene, the model may generate the wrong description. This problem can be well solved by increasing the number and variety of the dataset.
\par Generally speaking, two main reasons make the proposed model in this paper able to complete the image caption generation task very well. Firstly, the most advanced language model---GF-LSTM---is used in the proposed 3G model. Secondly, gated global image feature and attention-based local image feature are fused for image representation, which is beneficial to seize the accurate, comprehensive and meticulous information of images.
\section{Conclusion} \label{conclusion}
In this paper, a 3G model for image caption generation is proposed. The proposed model shows the better performance than other state-of-art model on three benchmark datasets. 3G model mainly has three gating structures: 1) gate for the global image feature, 2) gate for recurrent neural network and 3) gated feedback for multi-layer recurrent neural networks. Through the gated structure, the global image feature can be robustly selected to input into the multi-modal embedding model. We choose the gated recurrent neural network as language model because it solves the long term dependency problem in ``vanilla'' RNN. Gated feedback collection for multi-layer recurrent neural networks can deal with the problem of learning multiple adaptive timescales, and this makes GF-RNN more proper for language model than the standard stacked RNN. So, for both the image feature information and the language model, the proposed 3G model in this paper reinforces the recent state-of-the-art models for image caption generation.

\section*{Acknowledgement}
This work was supported in part by the National Natural Science Foundation of China under Grant nos. 61761130079, 61472413, and 61772510, in part by the Key Research Program of Frontier Sci- ences, CAS, under Grant no. QYZDY-SSW-JSC044, in part by the Young Top-Notch Talent Program of Chinese Academy of Sciences under Grant no. QYZDB-SSWJSC015 and in part by the National Key R\&D Program of China no. 2017YFB0502900.
\section*{References}
\bibliography{3G_NEROCOMPUTING}

%
%
%
%
\end{document}